\documentclass[runningheads]{llncs}
\usepackage{graphicx}
\usepackage{subfigure}
\usepackage{amsmath,amssymb} 
\usepackage{color}
\usepackage[width=122mm,left=12mm,paperwidth=146mm,height=193mm,top=12mm,paperheight=217mm]{geometry}
\begin{document}
\pagestyle{headings}
\mainmatter
\def\ECCV16SubNumber{***}  

\title{Analytical Derivatives for Differentiable~Renderer: 3D Pose Estimation by Silhouette~Consistency } 



\author{Zaiqiang Wu, Wei Jiang*}
\institute{Zhejiang University}

\maketitle

\begin{abstract}
Differentiable render is widely used in optimization-based 3D reconstruction which requires gradients from differentiable operations for gradient-based optimization. The existing differentiable renderers obtain the gradients of rendering via numerical technique which is of low accuracy and efficiency. Motivated by this fact, a differentiable mesh renderer with analytical gradients is proposed. The main obstacle of rasterization based rendering being differentiable is the discrete sampling operation. To make the rasterization differentiable, the pixel intensity is defined as a double integral over the pixel area and the integral is approximated by anti-aliasing with an average filter. Then the analytical gradients with respect to the vertices coordinates can be derived from the continuous definition of pixel intensity. To demonstrate the effectiveness and efficiency of the proposed differentiable renderer, experiments of 3D pose estimation by only multi-viewpoint silhouettes were conducted. The experimental results show that 3D pose estimation without 3D and 2D joints supervision is capable of producing competitive results both qualitatively and quantitatively. The experimental results also show that the proposed differentiable renderer is of higher accuracy and efficiency compared with previous method of differentiable renderer.
\end{abstract}

\section{Introduction}

In recent years, convolutional neural networks (CNNs) have achieved appealing results in image understanding, such as single image based 3D reconstruction. It is generally known that differentiable operations are essential for back-propagation algorithm to train the neural networks. For instance, 3D reconstruction in generative manner requires differentiable renderer to construct the loss for supervision. However, due to the discrete sampling operation, the traditional rendering algorithms, (e.g., rasterization and ray tracing~\cite{whitted2005improved}) are not differentiable and can not be directly applied in the framework of 3D reconstruction.

Many researchers paid a lot of attention to differentiate the process of rendering to make it feasible to incorporate the rendering operation into gradient-based optimization framework. Loper \textit{et~al.}~\cite{loper2014opendr} proposed a general-purpose differentiable renderer named OpenDR which is capable of rendering triangular meshes into images and automatically acquiring derivatives with respect to the model parameters. However the derivatives of OpenDR are computed by numerical method which is lack of accuracy. Further more, OpenDR is not compatible with existing deep learning framework. Kato \textit{et~al.}~\cite{kato2018neural} proposed a differentiable renderer designed for neural networks, but this method still relies on numerical methods to compute derivatives. Liu \textit{et al.}~\cite{liu2019soft} proposed a differentiable renderer called SoftRas which only focuses on the rendering of silhouette, however this method requires to generate probability maps for each triangle in the mesh, which results in high memory consumption and blurry rendering results.

To address these issues mentioned above, the first differentiable silhouette renderer with analytical derivatives which is of higher efficiency and accuracy compared with previous methods. It is worth mentioning that it is not necessary to utilize a general-purpose renderer in 3D reconstruction tasks since the illumination and material parameters are usually unknown, thus a differentiable renderer focusing on synthesizing silhouettes is enough for supervision. The forward pass of our renderer is similar to rasterization with anti-aliasing. However the backward pass is different from previous methods which depend on accessing to rendered frame buffers and obtaining derivatives by numerical methods. The high light of our work is that the derivatives of pixel intensities with respect to the coordinates of vertices are obtained by our proposed analytical method without the need of accessing to the frame buffers and applying any numerical method.

To obtain the derivatives of rasterization, the pixel intensities are defined as the average value of the certain area within the pixel region. The average value can be obtained by a double integral over pixel region of the pixel intensity function. Since only silhouette is considered in this paper, there is no need to deal with self-occlusion. Based on the integral expression of pixel intensity, the expression of derivatives could be obtained and simplified to an analytical expression without integral forms. With the analytical expression of derivatives, it is convenient and efficient to implement the backward pass of rendering. Our main contributions are summarized below.

\begin{itemize}
  \item The analytical expressions of derivatives of rasterization are derived and a novel non-numerical approach is proposed to implement the backward pass of differentiable renderer efficiently.
  \item Experiments were conducted to demonstrated that our proposed method is of higher accuracy and efficiency compared with previous state-of-the-art method.
  \item The potential of 3D pose estimation by silhouette consistency without 2D and 3D joints is shown in the experiments we conducted.
\end{itemize}

\section{Related Work}
\subsection{Differentiable Renderer}
Computer vision problems have been viewed as inverse graphics in a long literature. Computer graphics aims to render an image from the object shape, texture and illumination. In contrary to computer graphics, inverse graphics aims to estimate the object shape, texture and illumination from an input image. Differentiable rendering offers a straightforward and practical technique to infer the parameters of 3D models by gradient-based methods.

Gkioulekas \textit{et al.}~\cite{gkioulekas2016evaluation} developed an algorithmic framework to infer internal scattering parameters for heterogeneous materials. Gradients are leveraged for optimization to solve this inverse problem, however this approach is limited to specific illumination problems. Mansinghka \textit{et al.}~\cite{mansinghka2013approximate} proposed a probabilistic graphics model to estimate scene parameters from observations. Loper and Black~\cite{loper2014opendr} introduced an approximate differentiable renderer called OpenDR that makes it easy to render 3D model and automatically obtain derivatives w.r.t. the model parameters. However OpenDR has no interfaces to popular deep learning library which makes it difficult to be incorporated into deep learning framework. Kato \textit{et al.}~\cite{kato2018neural} introduced a differentiable rendering pipeline which approximate the rasterization gradient with a hand-designed function. More recently, Li \textit{et al.}~\cite{li2018differentiable} presented a differentiable ray tracer which is able to compute derivatives of scalar function over the rendered image w.r.t. arbitrary scene parameters. However the forward pass and backward pass of this method are performed by Monte Carlo ray tracing which makes it time consuming and impractical to be incorporated into learning-based framework.

With the development of deep learning and CNNs, there is a growing trend for researchers to achieve the froward pass and backward pass of differentiable rendering in a deep learning framework~\cite{zienkiewicz2016real,liu2017material,richardson2017learning,tewari2017mofa,tewari2018self,deschaintre2018single,kundu20183d,genova2018unsupervised}. Nguyen-Phuoc \textit{et al.}~\cite{nguyen2018rendernet} presented RenderNet, a convolutional network which learns the direct map from scene parameters to corresponding rendered images. However the shortcoming of RenderNet is that it is computational expensive since it is composed of convolutional networks.

In this paper, we focus on exploring a rasterization-based differentiable renderer with analytical derivatives. The main difference between our work and Neural 3D Mesh Render~\cite{kato2018neural} is that instead of approximating the derivatives with hand-designed functions we derived a analytical expression to obtain derivatives with significantly higher efficiency and accuracy.
\subsection{Single-image 3D reconstruction}
Inferring 3D shape from images is a traditional and challenging problem in computer vision. With the surge of deep learning, 3D reconstruction from a single image has become an active research topic in recent years.

Most of learning-based approaches learn the mapping from 2D image to 3D shape with 3D supervision. Some of these methods predict a depth map to reconstruct 3D shape~\cite{eigen2014depth,saxena20083}, while others predict 3D shapes directly~\cite{kato2018neural,wang2018pixel2mesh,choy20163d,fan2017point,tatarchenko2017octree,tulsiani2017multi,wu2016learning}.

When it comes to 3D pose estimation, statistical body shape models such as SMPL~\cite{loper2015smpl} and SCAPE~\cite{anguelov2005scape} are frequently employed due to their low dimensional representation. Bogo \textit{et al.}~\cite{bogo2016keep} proposed a iteratively optimization-based approach to reconstruct 3D human pose and shape from single image by minimizing the reprojection error between the 2D image and the statistical body shape model. Pavlakos \textit{et al.}~\cite{pavlakos2018learning} presented an end-to-end framework to predict the parameters of the statistical body shape model by training CNNs with single image and 3D ground truth.

Since 3D ground truth models are hard to obtain, 3D reconstruction without 3D supervision also attracts increasing attention. Yan \textit{et al.}~\cite{yan2016perspective} proposed perspective transformer nets (PTN) to infer 3D voxels from silhouette images from multiple viewpoints. Recent works predict 3D polygon meshes using differentiable renderer with 2D silhouettes supervision only. We follow these works in supervision, but we use a statistical body shape model named SMPL to represent 3D shape of human body and optimize the 3D pose with the gradients obtained by our proposed differentiable renderer.

\section{Analytical derivatives for rasterization}
Rasterization is a process of computing the mapping from scene geometry described in vector graphics format to raster images. The main obstacle that impedes rasterization from being differentiable is the discrete sampling operation that pixel intensities are sampled only at the central points of each pixel. Due to the discrete sampling operation and limited resolution, aliasing effect often appears in the rendered images. Anti-aliasing techniques are proposed to remove the aliasing effect and smooth the rendered images. In traditional anti-aliasing techniques, an image with higher resolution is rendered and down-sampled to the expected resolution with a average filter. Inspired by this approach, it is natural for us to assume that if a 3D model is rendered into an image with infinite resolution and down-sampled to the expected resolution using average filter, the sampling operation will be continuous and derivable. Since infinite resolution can not be achieved, the resolution of rendering is set to a higher and finite value to approximate the ideal situation. the forward pass of our renderer works the same as standard graphics pipeline with anti-aliasing but the backward derivatives are derived under the hypothesis that the image is rendered in infinite resolution and down-sampled into expected resolution by average filter.

\subsection{Forward rendering}
The forward pass of our proposed differentiable renderer follows the standard graphics method~\cite{marschner2015fundamentals}. To ensure the consistency between the forward and backward propagation, anti-aliasing is applied to smooth the rendered images.

Rendering a model with infinite resolution and down-sampling to the expected resolution, i.e., the pixel intensities equal to the double integral of a scalar function $p(x,y)$ of two variables $x$ and $y$ over the region within the pixel. The scalar function $p(x,y)$ represents the continuous distribution of intensity in the screen space.

Since we only focus on synthesizing silhouettes, i.e., there are only two possible values for $p(x,y)$: the foreground intensity $p_1$ and the background intensity $p_0$. Consider a image with $H$ rows and $W$ columns, the pixel intensity $I(i,j)$ of pixel in $i$-th row and $j$-th column can be represented as:
\begin{align}
\label{eq:continuous}
  I(i,j)=\frac{1}{S}\iint_{\Omega_{i,j}}p(x,y)\,dx\,dy
\end{align}
where $\Omega_{i,j}$ represents the region of the pixel in row $i$ and column $j$, $S$ denotes the area of region within the pixel.

\begin{figure}
\centering
\subfigure[Input mesh]{
\begin{minipage}[t]{0.3\linewidth}
\centering
\includegraphics[width=3.5cm]{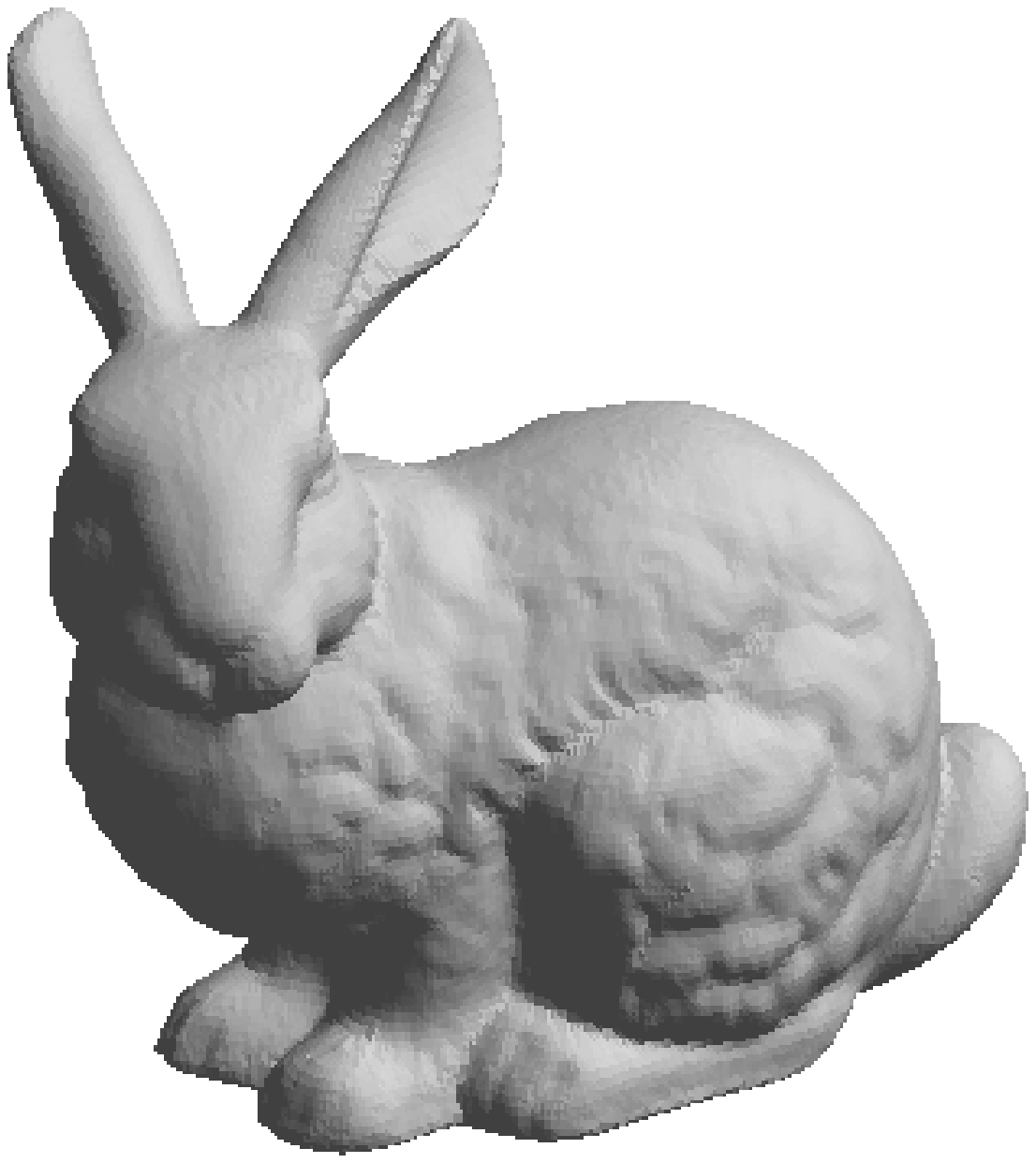}
\label{fig:forward_render:a}
\end{minipage}%
}
\subfigure[Before anti-aliasing]{
\begin{minipage}[t]{0.3\linewidth}
\centering
\includegraphics[width=3.5cm]{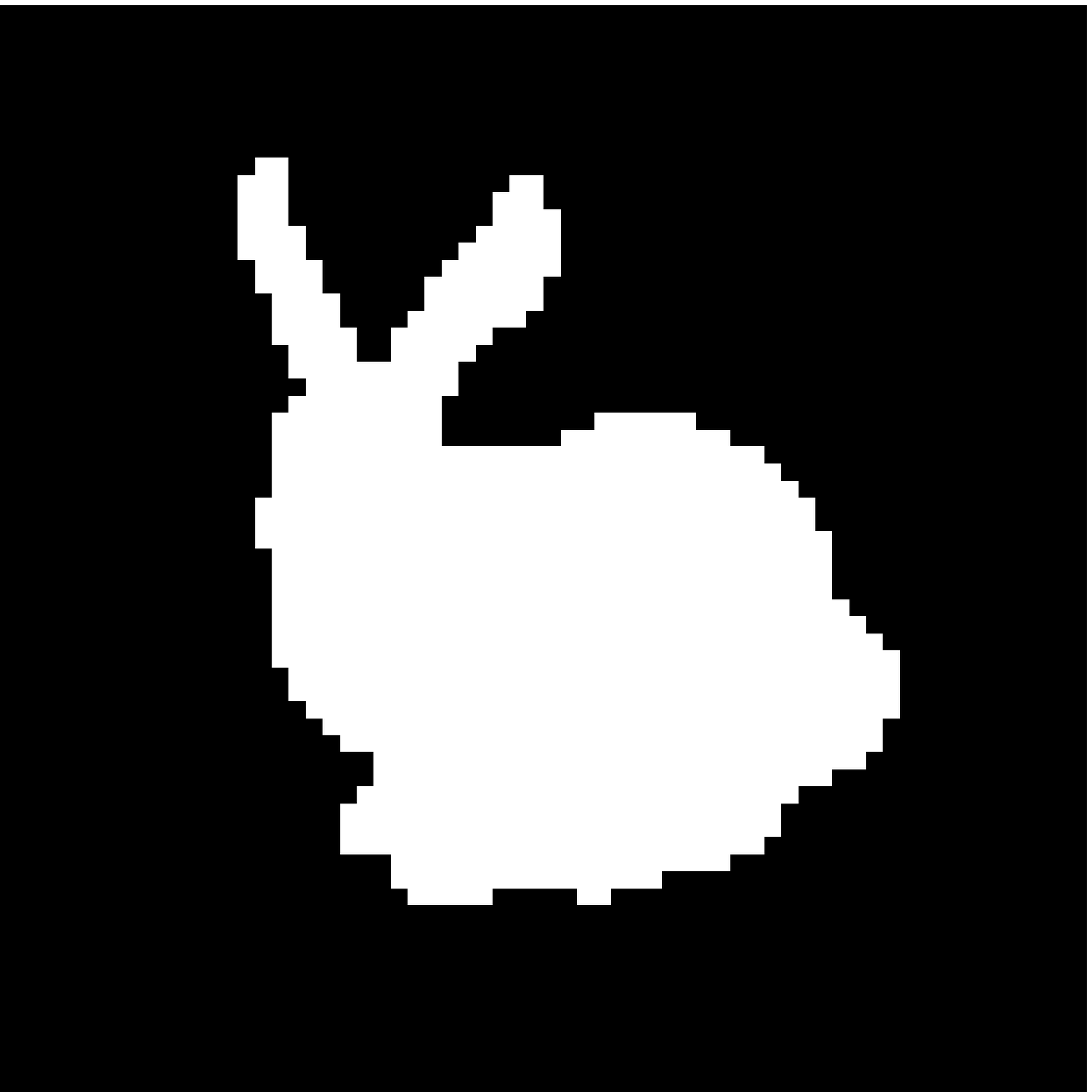}
\label{fig:forward_render:b}
\end{minipage}%
}
\subfigure[After anti-aliasing]{
\begin{minipage}[t]{0.3\linewidth}
\centering
\includegraphics[width=3.5cm]{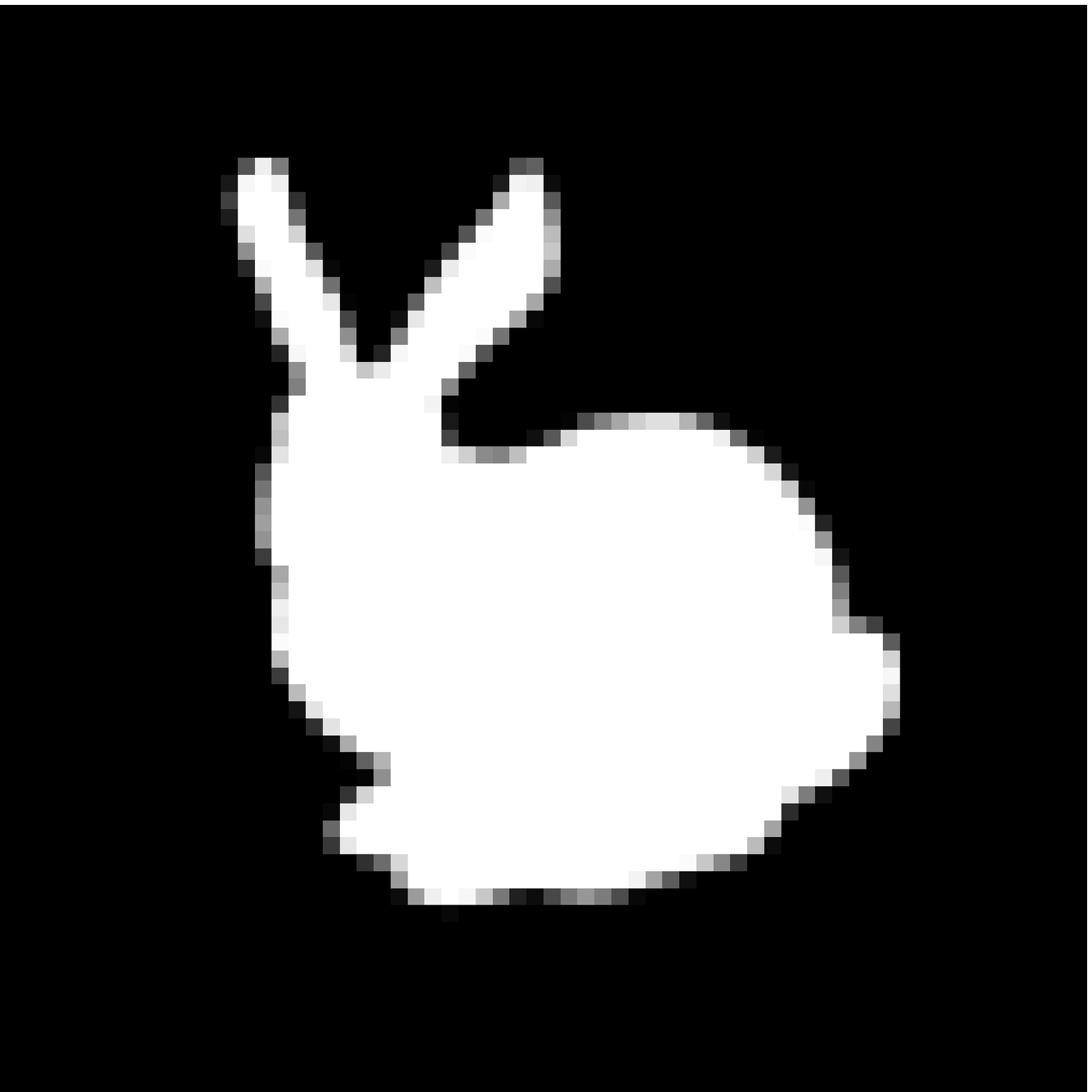}

\label{fig:forward_render:c}
\end{minipage}%
}
\caption{The forward rendering process of our differentiable renderer. To make the rendered silhouette image more smooth, we first render a silhouette image with higher resolution then down-sample it to get the final image.}
\label{fig:forward_render}
\end{figure}

However the value of the integral expression in Equation \ref{eq:continuous} is hard to compute in computer, so we use anti-aliasing to approximate this integral value as shown in Figure \ref{fig:forward_render}. The anti-aliasing we adopt is fairly rudimentary compared to more modern techniques. With this approach, individual pixels are divided into multiple coverage samples. By analyzing the intensity of the pixels surrounding each of these samples, an average intensity is produced, which determines the intensity of the original pixel. $F$ times anti-aliasing is applied in rendering, then the pixel intensity can be obtained as:
\begin{align}
  I(i,j)=\frac{1}{F^2}\sum_{k=1}^{F^2}p(x_k,y_k)
\end{align}
where $x_k$ and $y_k$ represent the coordinate of the $k$-th sampling point in screen space.

It is obvious that:
\begin{align}
  \lim_{F\to \infty}\frac{1}{F^2}\sum_{k=1}^{F^2}p(x_k,y_k)=\frac{1}{S}\iint_{\Omega_{i,j}}p(x,y)\,dx\,dy
\end{align}

When implementing the code, we set $F$ to $4$ for the tradeoff between accuracy and speed.

\subsection{Derivatives computation}
With the continuous definition of pixel intensity in Equation \ref{eq:continuous}, the derivatives with respect to the vertices can be derived. Considering a edge consisted of vertices $v_a$ and $v_b$ located at the boundary of the silhouette, the coordinates of $v_a$ and $v_b$ are denoted as $(x_0,y_0)$ and $(x_1,y_1)$. Assuming that this edge is intersected with the region of pixel in $i$-th row and $j$-th column. The partial derivative $\frac{\partial I(i,j)}{\partial x_0}$ can be written as:
\begin{align}
  \frac{\partial I(i,j)}{\partial x_0}&=\frac{\partial \frac{1}{S}\iint_{\Omega_{i,j}}p(x,y)\,dx\,dy}{\partial x_0}\\
  &=\frac{1}{S}\iint_{\Omega_{i,j}}\frac {\partial p(x,y)}{\partial x_0}\,dx\,dy
\end{align}

For notational convenience we denote that $A=y_1-y_0$, $B=x_0-x_1$, $C=x_1y_0-x_0y_1$. The equation of the edge can be represented as:
\begin{align}
  \alpha(x,y)=Ax+By+C \label{eq:line}
\end{align}

Assuming that if $\alpha(x,y)<0$, then the point $(x,y)$ is in the region of foreground, and vice versa. Let $\Omega_0$ be a appropriate sub region of $\Omega_{i,j}$ s.t. $\Omega_0$ only covers the edge connecting $v_a$ and $v_b$, thus the intensity distribution function $p(x,y)$ can be written as:
\begin{align}
  p(x,y)=
  \begin{cases}
p_1,  & \mbox{if }\alpha(x,y)<0 \mbox{ and } (x,y)\in\Omega_0 \\
p_0, & \mbox{if }\alpha(x,y)>0 \mbox{ and } (x,y)\in\Omega_0
\end{cases}
\end{align}

The equation above can be simplified with the Heaviside step function $h$:
\begin{align}
p(x,y)=p_0h(\alpha(x,y))+p_1h(-\alpha(x,y)), (x,y)\in\Omega_0 \label{eq:step}
\end{align}

The partial derivative $\frac{\partial I(i,j)}{\partial x_0}$ can be rewritten as:
\begin{align}
\frac{\partial I(i,j)}{\partial x_0}&=\frac{1}{S}\iint_{\Omega_0}\frac {\partial p(x,y)}{\partial x_0}\,dx\,dy + \frac{1}{S}\iint_{\Omega_{i,j}-\Omega_0}\frac {\partial p(x,y)}{\partial x_0}\,dx\,dy \\
&=\frac{1}{S}\iint_{\Omega_0}\frac {\partial p(x,y)}{\partial x_0}\,dx\,dy \label{eq:partial}
\end{align}

From Equation \ref{eq:step} and Equation \ref{eq:partial} we can obtain the partial derivative $\frac{\partial I(i,j)}{\partial x_0}$ as:
\begin{align}
\frac{\partial I(i,j)}{\partial x_0}&=\frac{1}{S}\iint_{\Omega_0}p_0\delta(\alpha(x,y))\frac{\partial \alpha(x,y)}{\partial x_0} -p_1\delta(\alpha(x,y))\frac{\partial \alpha(x,y)}{\partial x_0}\,dx\,dy \\
&=\frac{p_1-p_0}{S}\iint_{\Omega_0}\delta(\alpha(x,y))(-\frac{\partial \alpha(x,y)}{\partial x_0})\,dx\,dy \label{eq:derivative}
\end{align}
where $\delta$ denotes the Dirac delta function.

Substituting Equation \ref{eq:line} into Equation \ref{eq:derivative}, the partial derivative $\frac{\partial I(i,j)}{\partial x_0}$ can be represented as:
\begin{align}
\frac{\partial I(i,j)}{\partial x_0}=\frac{p_1-p_0}{S}\iint_{\Omega_0}\delta(Ax+By+C)(y_1-y)\,dx\,dy \label{eq:derivative1}
\end{align}

To eliminate the Dirac delta function, we perform the following variable substitution:
\begin{align}
\begin{cases}
t=Ax+By \\
k=-Bx+Ay
\end{cases}
\end{align}

After variable substitution, Equation \ref{eq:derivative1} can be rewritten as:
\begin{align}
\frac{\partial I(i,j)}{\partial x_0}&=\frac{p_1-p_0}{S(A^2+B^2)}\iint\delta(t+C)(y_1-\frac{Bt+Ak}{A^2+B^2})\,dt\,dk \\
&=\frac{p_1-p_0}{S(A^2+B^2)}\int_{k_0}^{k_1}(y_1-\frac{Ak-BC}{A^2+B^2})\,dk \\
&=\frac{p_1-p_0}{S(A^2+B^2)}((y_1+\frac{BC}{A^2+B^2})(k_1-k_0)-\frac{A(k_1^2-k_0^2)}{2(A^2+B^2)})
\end{align}
where $A^2+B^2$ is the $L^2$ length of the edge, which takes the Jacobian of the variable substitution into account. $k_0$ and $k_1$ are the lower and upper limits of integral obtained by Liang-Barsky algorithm~\cite{liang1984new}.

\begin{figure}
\centering
\includegraphics[width=8cm]{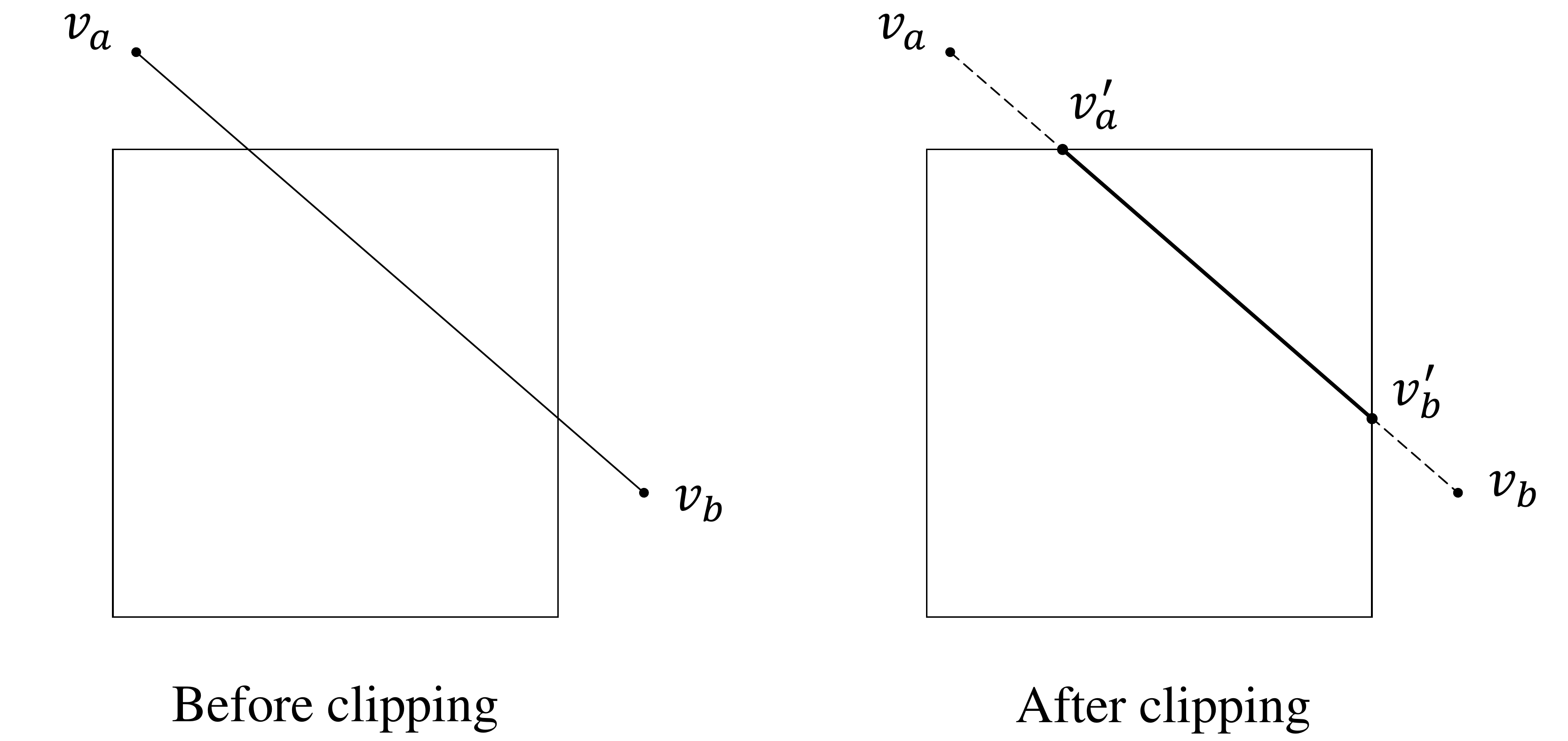}
\caption{Illustration of how to determine the lower and upper limits of the integral by Liang-Barsky algorithm. After clipping, two new endpoints $v_a^\prime$ and $v_b^\prime$ are obtained. Following the variable substitution, the coordinates of $v_a^\prime$ and $v_b^\prime$ can be transformed to the lower limit $k_0$ and upper limit $k_1$.}
\label{fig:Liang-Barsky}
\end{figure}

To illustrate the procedure of determining the lower and upper limits, the two new endpoints after clipping are denoted as $v_a^\prime$ and $v_b^\prime$ as shown in Figure \ref{fig:Liang-Barsky}, the coordinates are denoted as $(x_0^\prime,y_0^\prime)$ and $(x_1^\prime,y_1^\prime)$ respectively. Then the lower and upper limits can be obtained as:
\begin{align}
\begin{cases}
k_0&=-Bx_0^\prime+Ay_0^\prime \\
k_1&=-Bx_1^\prime+Ay_1^\prime
\end{cases}
\end{align}

The same procedure can be easily adapted to obtain the partial derivatives $\frac{\partial I(i,j)}{\partial y_0}$, $\frac{\partial I(i,j)}{\partial x_1}$ and $\frac{\partial I(i,j)}{\partial y_1}$ as follows.
\begin{align}
\frac{\partial I(i,j)}{\partial y_0}&=-\frac{p_1-p_0}{S(A^2+B^2)}((x_1+\frac{AC}{A^2+B^2})(k_1-k_0)+\frac{B(k_1^2-k_0^2)}{2(A^2+B^2)})\\
\frac{\partial I(i,j)}{\partial x_1}&=\frac{p_1-p_0}{S(A^2+B^2)}(-(y_0+\frac{BC}{A^2+B^2})(k_1-k_0)+\frac{A(k_1^2-k_0^2)}{2(A^2+B^2)})\\
\frac{\partial I(i,j)}{\partial y_1}&=\frac{p_1-p_0}{S(A^2+B^2)}((x_0+\frac{AC}{A^2+B^2})(k_1-k_0)+\frac{B(k_1^2-k_0^2)}{2(A^2+B^2)})
\end{align}

It is feasible to obtain the derivatives without any numerical method with the analytical expressions of derivatives above, which brings space for improvement in accuracy and efficiency.
\subsection{Backward gradients flow}
Considering a 3D mesh consisting of a set of vertices $\{v_1^o,v_2^o,\dots,v_{N_v}^0\}$ and faces $\{f_1,f_2,\dots,f_{N_f}\}$. $v_k^o\in\mathbb{R}^3$ represents the position of the $k$-th vertex in the 3D object space and $f_k\in\mathbb{N}^3$ represents the the indices of the three vertices corresponding to the $k$-th triangle face. For rendering this 3D mesh, vertices $\{v_k^o\}$ in the object space are projected into screen space as vertices $\{v_k\}, v_k\in\mathbb{R}^2$.

 The scalar loss function over the rendered image for optimization is denoted as $L$. The partial derivatives $\{\frac{\partial L}{\partial I(i,j)}|i=1,\dots,H,j=1,\dots,W\}$ can be computed through automatic differentiable library. Our task is that: given the partial derivatives of loss function $L$ with respect to pixel intensities $\{\frac{\partial L}{\partial I(i,j)}\}$, our goal is to compute derivatives of pixel intensities with respect to vertices $\{\frac{\partial I(i,j)}{\partial v_k}\}$. Thus the derivatives $\{\frac{\partial L}{\partial v_k}\}$ can be obtained by chain rule, after which the gradient backward flow will be completed.

 It should be noted that the gradients flow is sparse since $\frac{\partial I(i,j)}{\partial v_k}\ne0$ only if there is at least one edge consisted of $v_k$ intersected with the pixel region of $I(i,j)$. We only have to focus on specific $i$, $j$ and $k$ such that $\frac{\partial I(i,j)}{\partial v_k}\ne0$, this allows skipping pixels that have no contribution of gradient to current triangle when traversing the arrays of triangles and improves the efficiency.

\begin{figure}
\centering
\includegraphics[width=12cm]{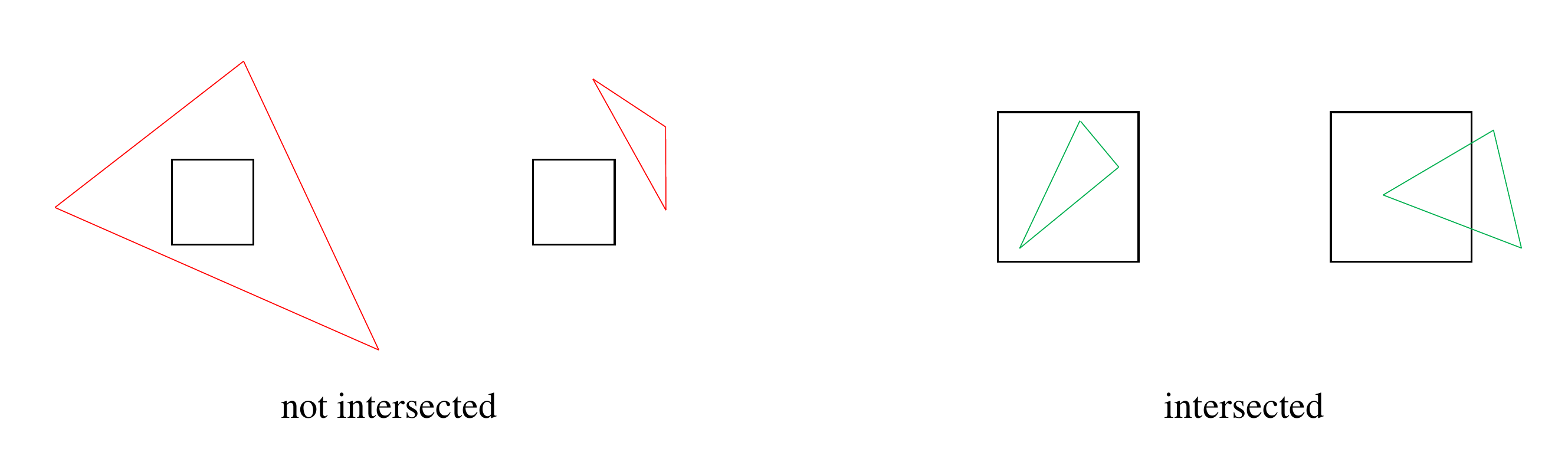}
\caption{Several intuitive examples of wether a triangle is intersected with the pixel.}
\label{fig:Quadrant}
\end{figure}

In order to achieve efficient retrieval of pixels that have contribution of gradient to current triangle, pixels out of the bounding box of current triangle are excluded first. The Liang-Barsky clipping algorithm~\cite{liang1984new} is adopted to determine wether a pixel is intersected with current triangle. As shown in Figure \ref{fig:Quadrant}, a pixel is intersected with the triangle only if there is at least one edge of the triangle intersected with the pixel.

It is obvious that gradients only flow at the boundary pixel of the silhouette image, so edge detection is performed on the rendered image to determine pixels that gradients can flow into, computation is required only at the boundary of silhouette.

Considering a pixel at the boundary and it is in the $i$-th row and $j$-th column, we need to determine the partial derivatives of pixel intensity with respect to the location of $k$-th vertices $v_k$, denoted as $\frac{\partial I(i,j)}{\partial v_k}$. It is assumed that there are $N_e$ edges consisted of $v_k$ intersected with the pixel in row $i$, column $j$. The derivatives of the pixel intensity $I(i,j)$ with respect to the position of $v_k$ can be represented as:
\begin{align}
\frac{\partial I(i,j)}{\partial v_k}=
\begin{cases}
\sum_{n=1}^{N_e}\frac{\partial I(i,j)}{\partial v_k^n}, & \mbox{if }N_e>0 \\
0, & \mbox{if }N_e=0
\end{cases}
\end{align}
where $\frac{\partial I(i,j)}{\partial v_k^n}$ represents the derivatives computed by the $n$-th edge.

\begin{figure}
\centering
\subfigure[Moving right]{
\begin{minipage}[t]{0.2\linewidth}
\centering
\includegraphics[width=2.5cm]{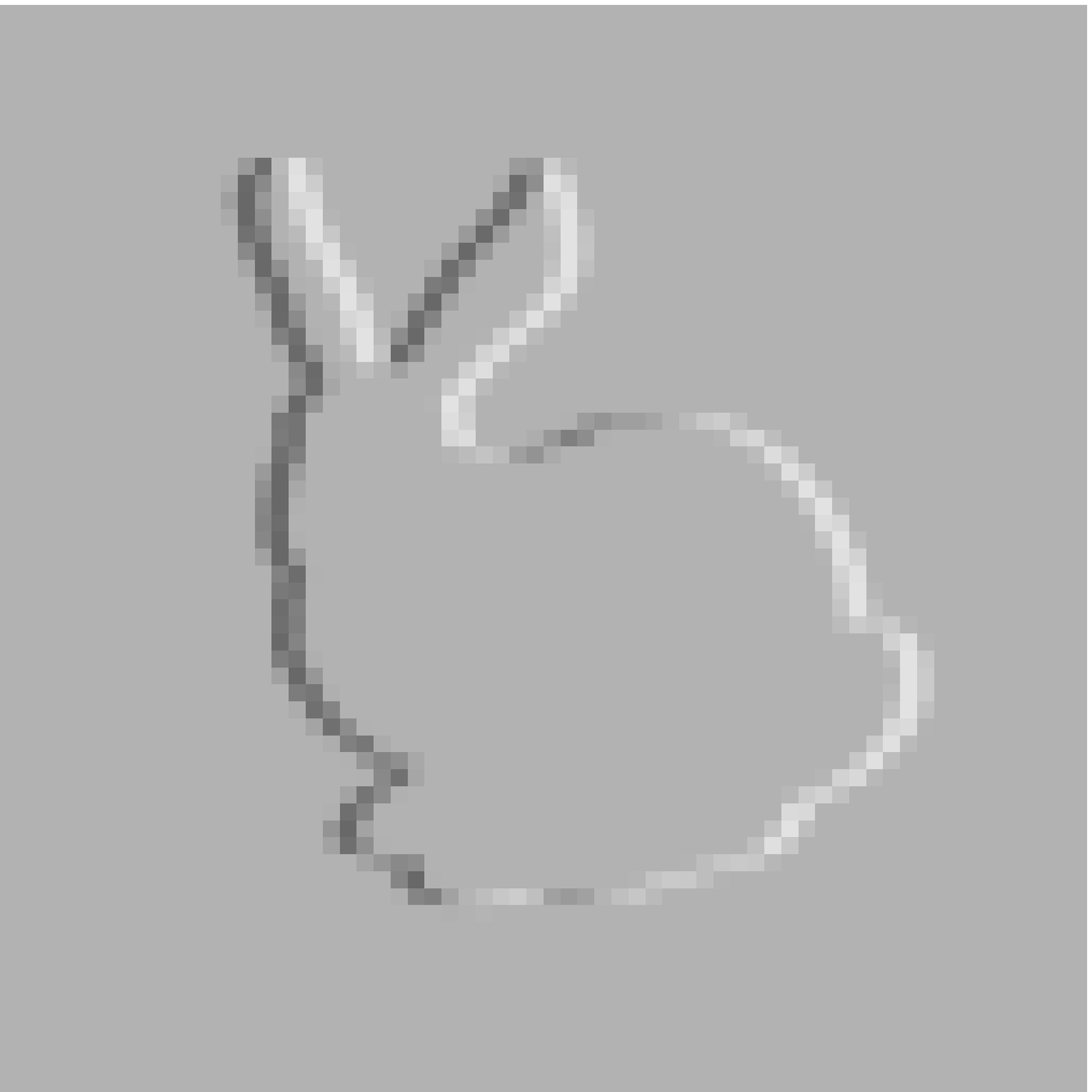}
\label{fig:gradient_map:a}
\end{minipage}%
}
\subfigure[Rotating]{
\begin{minipage}[t]{0.2\linewidth}
\centering
\includegraphics[width=2.5cm]{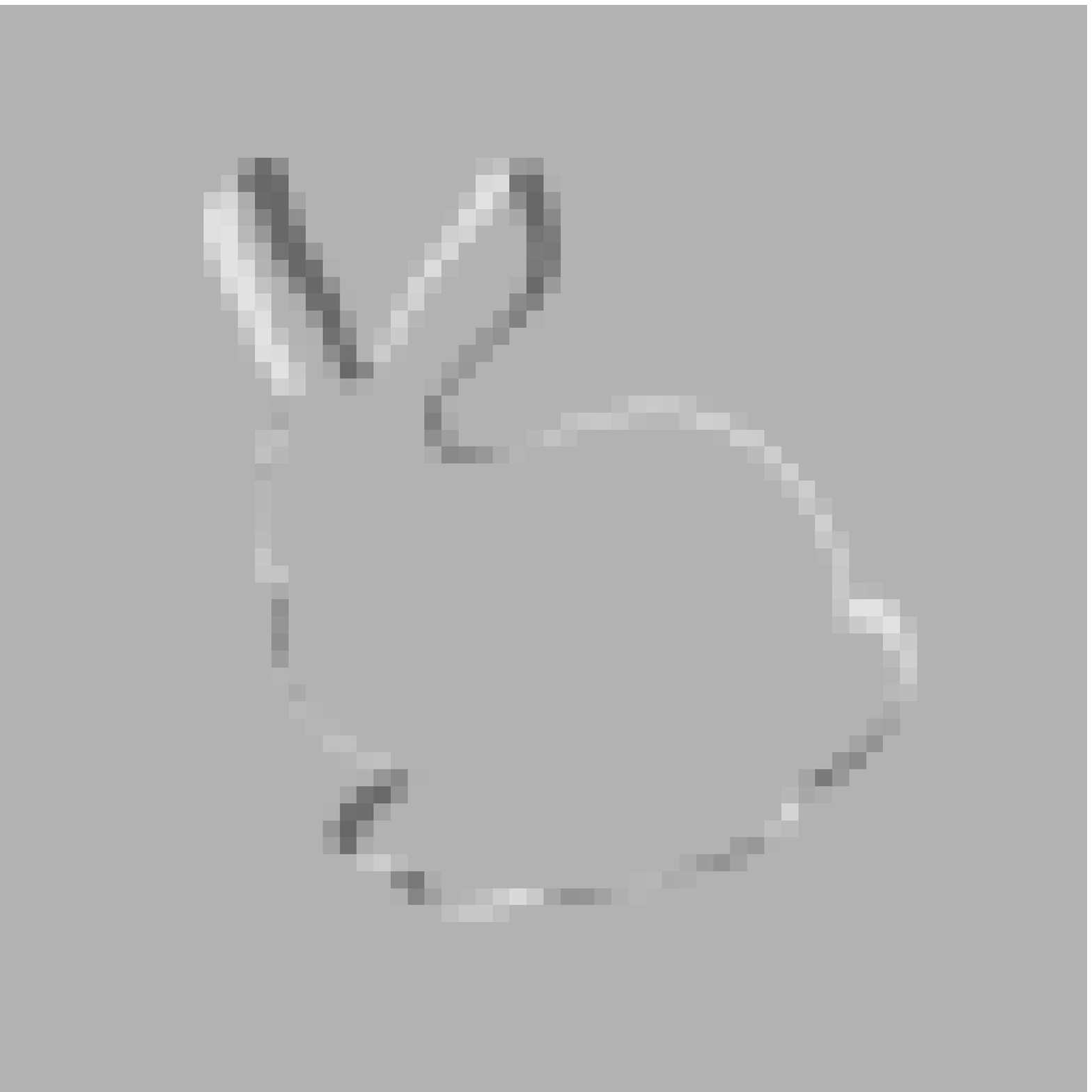}
\label{fig:gradient_map:b}
\end{minipage}%
}
\subfigure[Scaling up]{
\begin{minipage}[t]{0.2\linewidth}
\centering
\includegraphics[width=2.5cm]{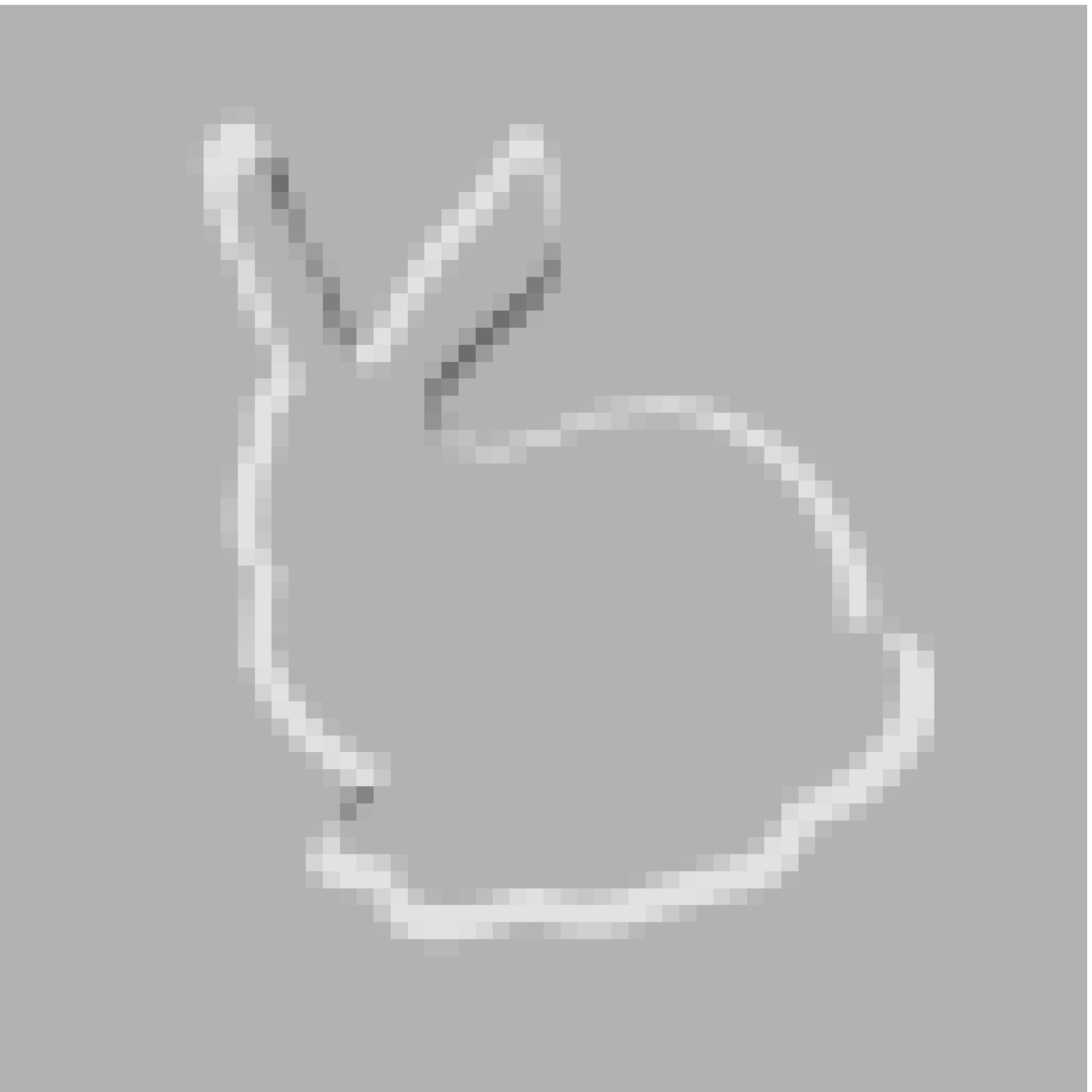}
\label{fig:gradient_map:c}
\end{minipage}%
}
\subfigure[Scaling down]{
\begin{minipage}[t]{0.2\linewidth}
\centering
\includegraphics[width=2.5cm]{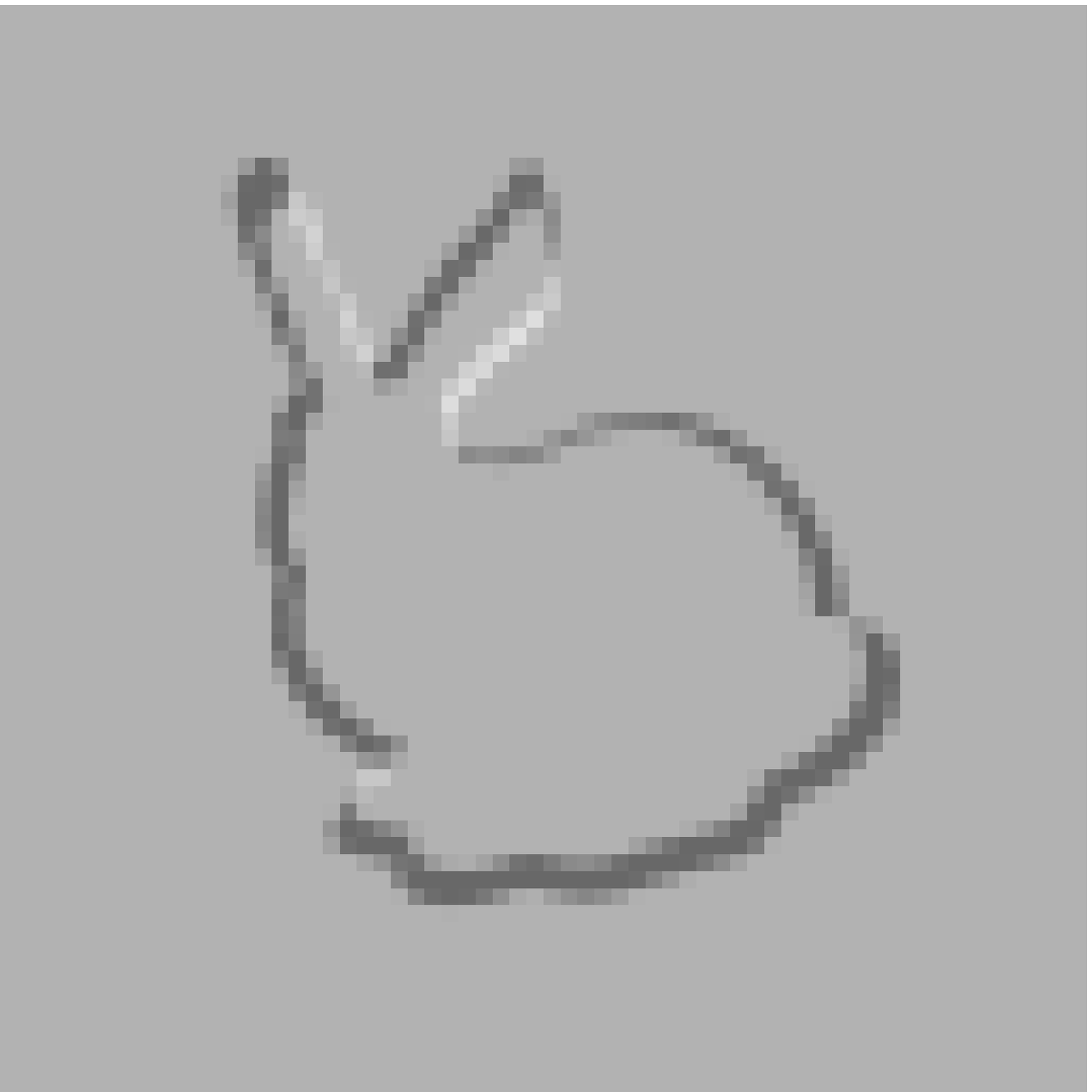}
\label{fig:gradient_map:d}
\end{minipage}%
}
\caption{Visualized per-pixel gradient with respect to different parameters obtained by our differentiable renderer. The gradients are with respect to (a) the bunny model moving right (b) the bunny model rotating anti-clockwise (c) the bunny model scaling up (d) the bunny model scaling down.}
\label{fig:gradient_map}
\end{figure}

To verify our method, experiments of our differentiable renderer on generating per-pixel gradient with respect to translation, rotation and scaling were conducted. The visualized results are presented in Figure \ref{fig:gradient_map}. From the visualized per-pixel gradient images, conclusion can be draw that our proposed differentiable renderer is able to generate correct gradients with respect to vertices location, which enables the gradient-based optimization for 3D pose estimation.
\section{3D pose estimation}
To show the effectiveness of our method, experiments of 3D pose estimation based on statistical body shape model by our proposed differentiable silhouette renderer were performed. Following the work of \cite{bogo2016keep}, an iteratively optimization-based method is presented to estimate the pose parameters of statistical body shape model by minimizing the error between reprojected silhouettes and ground truth silhouettes. The images and 3D ground truth leveraged in the experiments are from a 3D pose dataset named  UP-3D~\cite{lassner2017unite}. Unlike previous works, ground truth 2D and 3D joints truth are not necessary for experiments of 3D pose estimation in this paper.
\subsection{Statistical body shape model}
 A statistical body shape model named SMPL~\cite{loper2015smpl} is employed as our representation of 3D body model. Essential notations of SMPL model are provided here. The SMPL model can be view as s function $\mathcal{M}(\beta,\theta;\Phi)$, where $\beta$ is the shape parameters, $\theta$ is the pose parameters and $\Phi$ are fixed parameters learned from a dataset with body scans~\cite{robinette2002civilian}. The output of the SMPL function are vertices $P\in \mathbb{R}^{N\times3}$ with $N=6890$ of a body mesh. The shape parameters $\beta\in \mathbb{R}^{10}$ are the linear coefficients of a low number of principal body shapes. The pose parameters $\theta\in\mathbb{R}^{24\times3}$ are expressed in axis and angle representation and define the relative rotation between parts of the skeleton. Additionally, the 3D joints $J\in\mathbb{R}^{24\times3}$ obtained conveniently by a sparse linear combination of mesh vertices.

In our experiments, the shape parameters $\beta$ are fixed and our goal is optimizing the pose parameters $\theta$ to minimize the errors between the ground truth silhouettes and reprojected silhouettes.

\subsection{Data preparation}
\begin{figure}
\centering
\includegraphics[width=12cm]{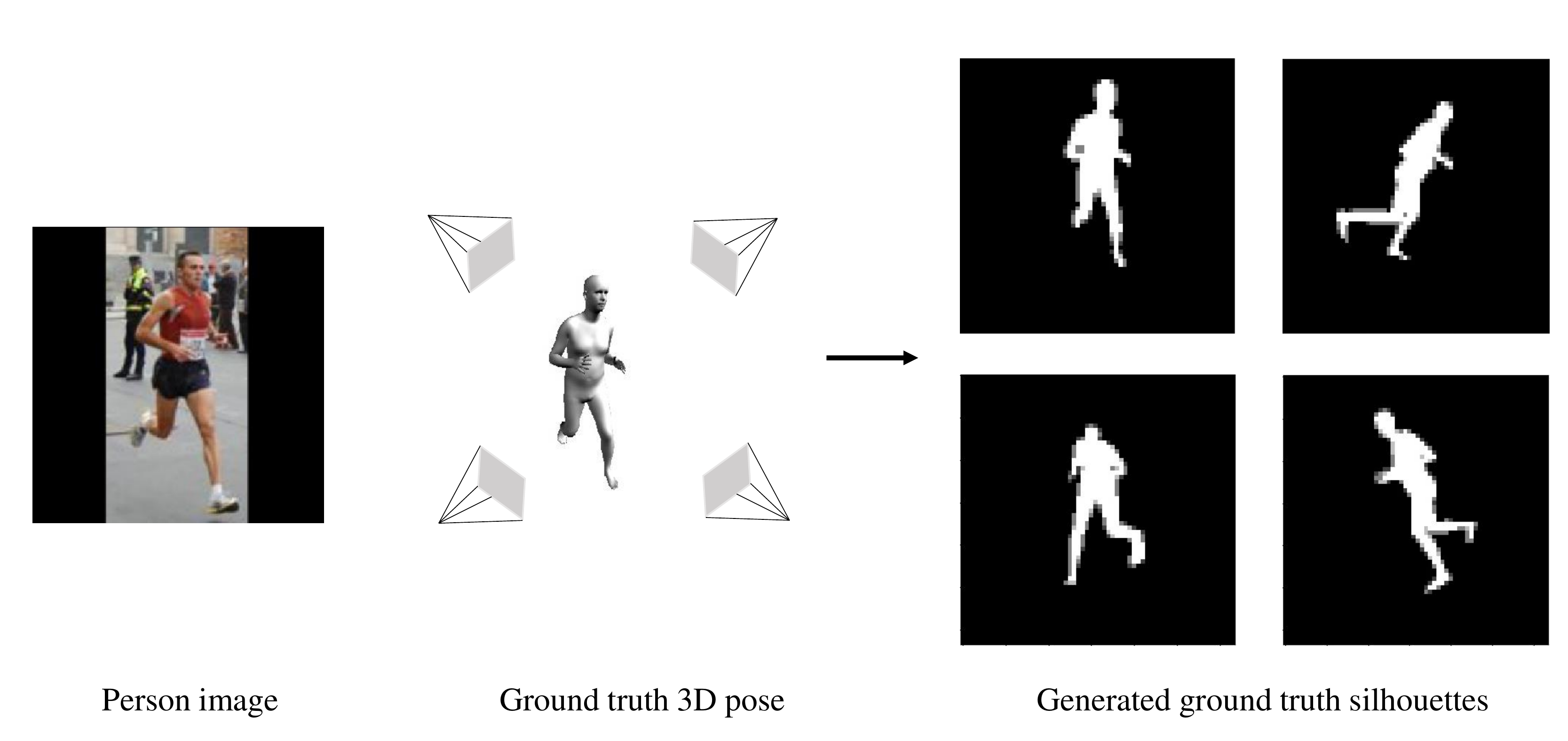}
\caption{The ground truth silhouettes for supervision are generated by projecting the ground truth 3D model to the image plane by cameras in different viewpoints.}
\label{fig:data_generation}
\end{figure}

 It is assumed that only images and multi-viewpoints silhouettes are available in the 3D pose estimation task. The ground truth silhouettes are generated by rendering the 3D ground truth meshes of UP-3D~\cite{lassner2017unite} from $4$ azimuth angles (with step of $90^\circ$) with fixed elevation angles ($0^\circ$) under the same camera setup as illustrated in Figure \ref{fig:data_generation}. The resolution of silhouettes is set to $64\times64$.

\subsection{Method}
Given a single image $I$ and its multi-viewpoints 2D silhouettes $\{S_i\}$, the 3D body model is fitted by minimizing a weighted sum of error terms.

The differentiable silhouette rendering process is denoted as $\mathcal{R}$, then the silhouette error term $E_{sl}$ can be represented as:
\begin{align}
E_{sl}&=\sum_{i=1}^{N_s}\lVert \mathcal{R}_i(\hat{P})-\mathcal{R}_i(P) \rVert_2^2\\
  &=\sum_{i=1}^{N_s}\lVert \mathcal{R}_i(\hat{P})-S_i \rVert_2^2 \\
  &=\sum_{i=1}^{N_s}\lVert \mathcal{R}_i(\mathcal{M}(\beta,\theta;\Phi))-S_i \rVert_2^2
\end{align}
where $P$ and $\hat{P}$ denote the ground truth vertices and estimated vertices, $N_s$ denotes the total number of silhouettes, $\mathcal{R}_i$ denotes the camera in the $i$-th position, $S_i$ denotes the $i$-th ground truth silhouette.

To discourage the body model from self-intersection, a self-intersection penalty term $E_{spt}$ from \cite{wu2019novel} is adopted. This self-intersection penalty term can be represented as:
\begin{align}
E_{spt}=\frac{N_{sec}}{N_v}
\end{align}
where $N_{sec}$ denotes the number of vertices in self-intersection region, $N_v$ denotes the total number of vertices.

The backward gradients of $E_{spt}$ is obtained by a hand-designed algorithm which can produce gradients to pull vertices out of region of self-intersection. The details of this algorithm are beyond the scope of this paper, we refer the interested readers to \cite{wu2019novel} for more details.

The objective function can be written as the weighted sum of the two error terms above:
\begin{align}
E=E_{sl}+\lambda E_{spt}
\label{eq:objective}
\end{align}
where $\lambda$ is a scalar weight.

\section{Experiments}
In this section, experiments of 3D pose estimation are performed to evaluate the effectiveness of our method. The details of our experiments setup are provided. The results of qualitative comparison and quantitative comparison are presented to demonstrate the effectiveness of our method.

\subsection{Experimental setup}
\subsubsection{Dataset}
Our proposed method is tested on UP-3D~\cite{lassner2017unite} for evaluation. This dataset contains color images and corresponding ground truth 3D pose represented as pose parameters of SMPL model. Noting that our iterative optimization-based method is sensitive to the initial pose, results on the subset of UP-3D selected by Tan \textit{et al.}~\cite{tan2018indirect} aiming to limit the range of global rotation of SMPL models are reported.
\subsubsection{Evaluation metric}

\begin{figure}
\centering
\includegraphics[width=8cm]{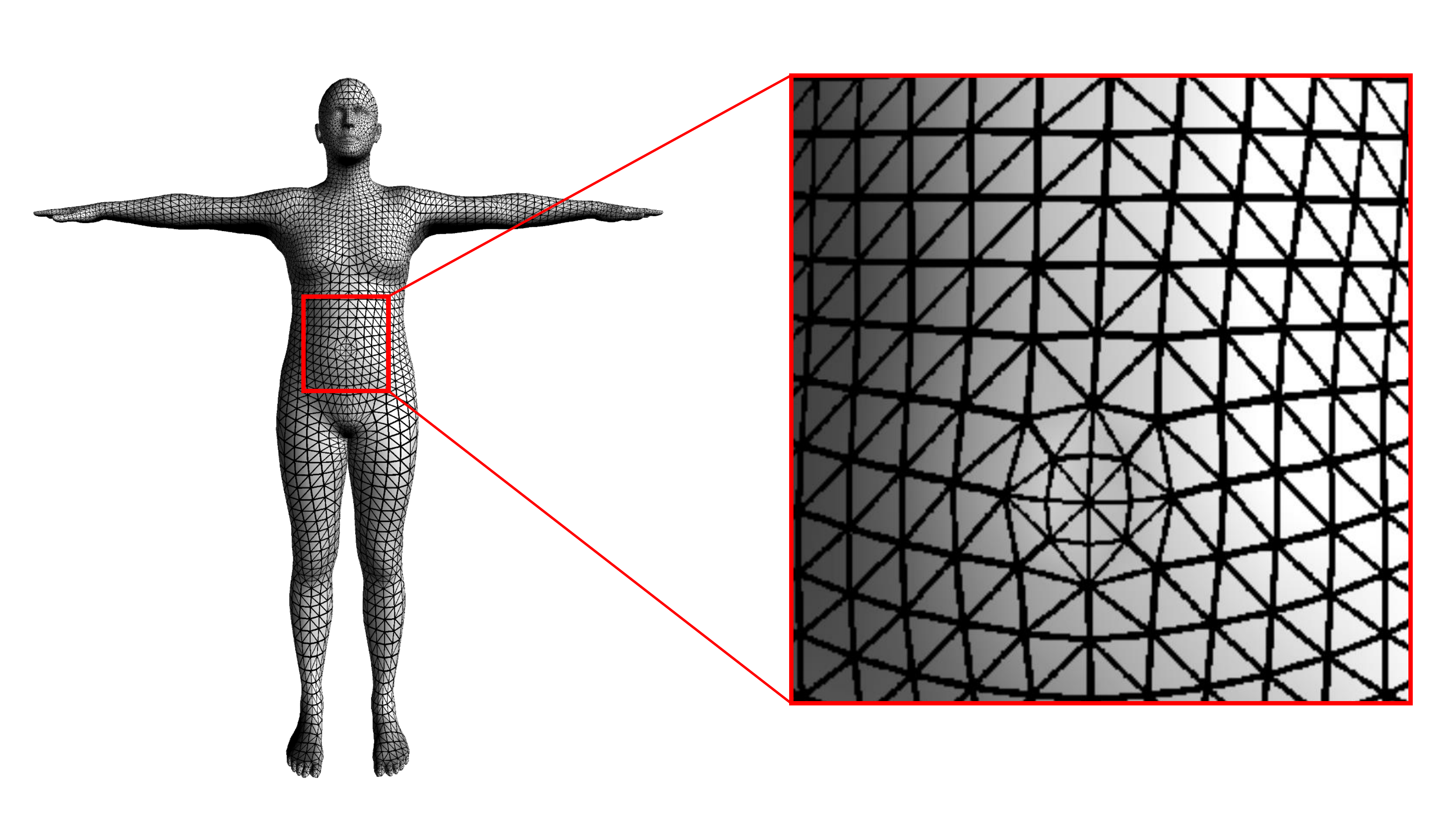}
\caption{Detail of SMPL mesh model. The SMPL mesh model is a vertex-based model that accurately represents body shapes by vertices and triangles.}
\label{fig:mesh}
\end{figure}

For quantitative evaluation, per-vertex error from~\cite{pavlakos2018learning} is used as metric for evaluating the accuracy of 3D pose when comparing with other methods. As shown in Figure \ref{fig:mesh}, the surface of body mesh is represented as vertices and triangles. The accuracy of pose estimation can be effectively evaluated by measuring error of each vertex, the per-vertex error $E_p$ can be represented as:
\begin{align}
E_p=\frac{1}{N_v}\sum_{i=1}^{N_v}\lVert \hat{P}_i-P_i\rVert_2
\end{align}
where $N_v$ denotes the total number of vertices, $\hat{P}_i$ denotes the estimated location of vertices, $P_i$ denotes the ground truth location of vertices.
\subsubsection{Implementation details}
The resolution of output images of differentiable renderer is set to $64\times64$, and the multiple of anti-aliasing $F$ is set to $4$. The number of silhouettes $N_s$ is set to $4$. The code is implemented in C++ with interface to the automatic differentiation library PyTorch~\cite{paszke2017automatic}, which allows us to employ their built-in optimizers and optimize the pose parameters of SMPL model easily. The objective function is minimized with Adam optimizer~\cite{kingma2014adam} with $\alpha=1.5\times10^{-4}$, $\beta_1=0.9$ and $\beta_2=0.999$. $\lambda$ in Equation \ref{eq:objective} is set to $0.001$ across all experiments.
\subsection{Qualitative comparison}
Comparison between the proposed differentiable renderer with Neural 3D Mesh Render (N3MR)~\cite{kato2018neural} is performed by conducting 3D pose estimation in same experimental setup. To demonstrate the effectiveness of our approach, we also compare our results with that of direct prediction method named Learning to Estimate 3D Human Pose and Shape from a Single Color Image (L2EPS) by Pavlakos~\textit{et al.} \cite{pavlakos2018learning}.

\begin{figure}
\centering
\includegraphics[width=12cm]{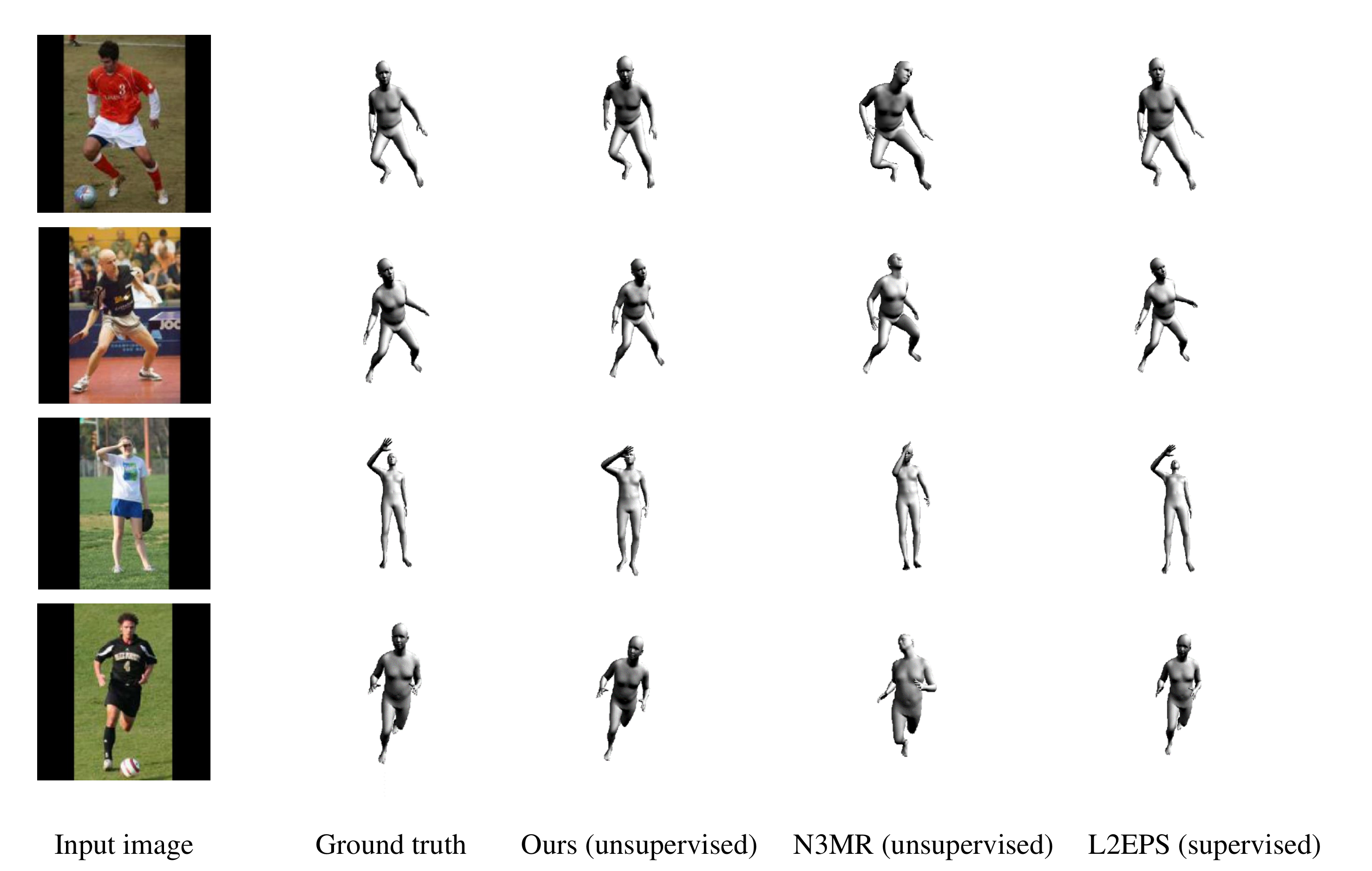}
\caption{Visualized results of 3D pose estimation by different methods. From left to right, we show the input images, ground truth, the results obtained by our method, the results obtained by N3MR~\cite{kato2018neural} and results of L2EPS~\cite{pavlakos2018learning}.}
\label{fig:qualitative}
\end{figure}

From the results shown in Figure \ref{fig:qualitative}, it is apparent that the Neural 3D Mesh Render suffers from local minimums which often result in failed prediction. Due to the discontinuous forward rendering pass without any smooth filter and the inconsistency between forward and backward propagations, the process of optimization is unstable and tends to fall in local minimums. In contrast, we apply anti-aliasing in the forward rendering to make the intensity of each pixel as much as possible close to the continuous definition in Equation \ref{eq:continuous}, which achieves the consistency between forward and backward propagations and stability of optimization.

Though our method performs 3D pose estimation without any 2D joint error term, the results are comparable with the learning-based method \cite{pavlakos2018learning} whose model is trained with 3D ground truth. Since 3D ground truth and 2D location are apparently more difficult to obtain than silhouette, our method offers possibility for 3D pose estimation without any 2D joint location and 3D ground truth.
\subsection{Quantitative comparison}
We show the quantitative evaluation on per vertex error with different approaches. Results are given in Table~\ref{table:quantitative}. As seen in Table~\ref{table:quantitative}, our differentiable renderer outperforms N3MR~\cite{kato2018neural} in 3D pose estimation. The result of our method is worse than that of L2EPS~\cite{pavlakos2018learning} since the method in \cite{pavlakos2018learning} leverages 3D ground truth but our method only leverages 2D silhouettes and predict 3D pose in an unsupervised manner.
\setlength{\tabcolsep}{4pt}
\begin{table}
\begin{center}
\caption{Quantitative results compared with other state-of-the-art methods}
\label{table:quantitative}
\begin{tabular}{lll}
\hline\noalign{\smallskip}
Method  & Per-vertex error (mm)\\
\noalign{\smallskip}
\hline
\noalign{\smallskip}
L2EPS~\cite{pavlakos2018learning} (supervised)  &117.7  \\
N3MR~\cite{kato2018neural} (unsupervised) &172.2 \\
Ours (unsupervised)&142.8 \\
\hline
\end{tabular}
\end{center}
\end{table}
\setlength{\tabcolsep}{1.4pt}
\subsection{Ablation analysis}
In this section, we conduct controlled experiments to validate the necessity of different components.
\subsubsection{Self-intersection penalty term.}
\begin{figure}
\centering
\includegraphics[width=11cm]{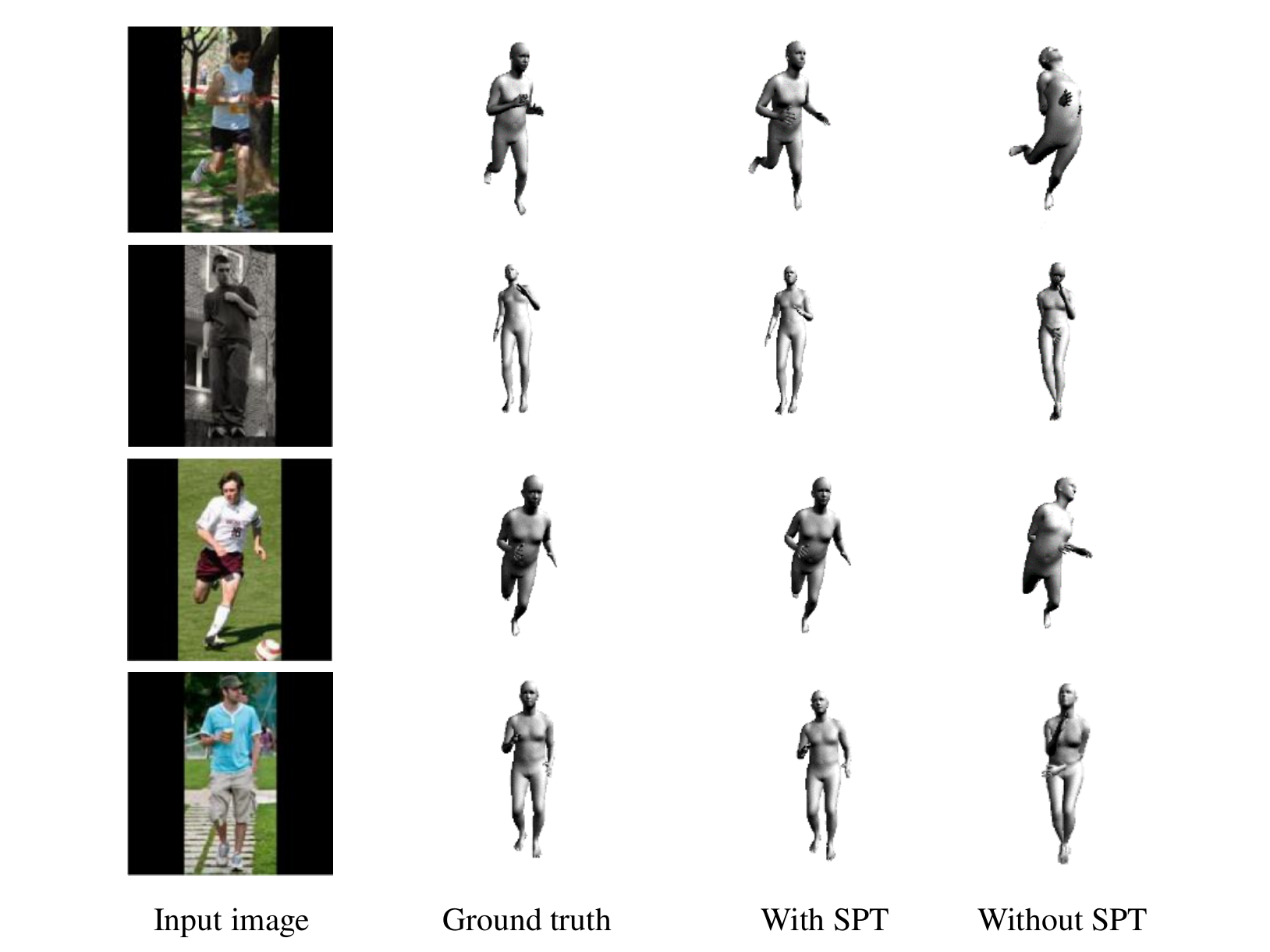}
\caption{Results of 3D pose estimation with and without SPT term. From left to right: input image, ground truth, prediction with SPT term and prediction without SPT term.}
\label{fig:self-intersection}
\end{figure}

We investigate the influence of Self-intersection penalty term in 3D pose estimation by conducting experiment without the self-intersection penalty term~\cite{wu2019novel} (SPT). In Figure \ref{fig:self-intersection} we visually compare the results of 3D pose estimation with and without SPT. As shown in Figure \ref{fig:self-intersection}, the result without SPT suffers from self-intersection. However the experiment with SPT obtains more reasonable result.
\subsubsection{Anti-aliasing.}
To demonstrate the importance of anti-aliasing in the forward pass of our differentiable renderer, we conduct quantitative comparison of 3D pose estimation by differentiable renderer with and without anti-aliasing. The result is given in Table \ref{table:anti-aliasing}, As seen in Table \ref{table:anti-aliasing}, anti-aliasing improves the accuracy of 3D pose estimation, especially when the resolution is quite low.
\setlength{\tabcolsep}{4pt}
\begin{table}
\begin{center}
\caption{Quantitative results of different rendering resolution and wether anti-aliasing is applied
caption should end without a full stop}
\label{table:anti-aliasing}
\begin{tabular}{lll}
\hline\noalign{\smallskip}
Resolution  & Anti-aliasing & Per-vertex error (mm)\\
\noalign{\smallskip}
\hline
\noalign{\smallskip}
$32\times32$  & No  & 181.9\\
$32\times32$ & Yes  &173.7\\
$64\times64$ & No  &153.4\\
$64\times64$ & Yes &142.8\\

\hline
\end{tabular}
\end{center}
\end{table}
\setlength{\tabcolsep}{1.4pt}
\subsection{Running time analysis}
To demonstrate the efficiency of our differentiable renderer, we carried out experiments of our method with different resolution and different number of SMPL models compared with N3MR~\cite{kato2018neural}. For a fair comparison, we implemented the CPU version of N3MR from their released GPU version. All experiments in this section were performed on a laptop with Intel(R) Core(TM) i7-8750H processer. We recorded the elapsed time of a single forward and backward pass of the two different renderer in Table \ref{table:running-time0} and Table \ref{table:running-time1}. As seen in Table \ref{table:running-time0} and Table \ref{table:running-time1}, with the increasing number of triangles and resolution, it is more and more obvious that our method runs faster than N3MR.
\setlength{\tabcolsep}{4pt}
\begin{table}
\begin{center}
\caption{Elapsed time in ms of one iteration of our method and N3MR in different resolution setup. The number of SMPL models is set to 1}
\label{table:running-time0}
\begin{tabular}{lll}
\hline\noalign{\smallskip}
Resolution  &  Ours  & N3MR \\

\noalign{\smallskip}
\hline
\noalign{\smallskip}
$16\times16$  & 17.11 &16.08 \\
$32\times32$ & 17.21 &16.11 \\
$64\times64$ & 17.63 &18.00\\
$128\times128$ & 18.50 & 23.64\\
\hline
\end{tabular}
\end{center}
\end{table}

\setlength{\tabcolsep}{4pt}
\begin{table}
\begin{center}
\caption{Elapsed time in ms of one iteration of our method and N3MR with different number of SMPL model. The rendering resolution  is set to $64\times64$}
\label{table:running-time1}
\begin{tabular}{lll}
\hline\noalign{\smallskip}
Number of SMPL  &  Ours  & N3MR \\

\noalign{\smallskip}
\hline
\noalign{\smallskip}
1  & 17.63 &18.00 \\
2 & 28.35 &29.20 \\
3 & 36.72 &38.96\\
4 & 46.65 & 49.14\\
\hline
\end{tabular}
\end{center}
\end{table}
\section{Conclusion}
In this paper, we proposed a novel method to obtain analytical derivatives for differentiable silhouette renderer. We demonstrate experiments of 3D pose estimation by silhouette consistency to show the effectiveness efficiency of our proposed method. Unlike pervious works like N3MR~\cite{kato2018neural} using numerical approach to obtain derivatives, we proposed a continuous definition of pixel intensity and derived the analytical derivatives based on the continuous definition. We adopt anti-aliasing to make sure the intensity of each pixel is close to the continuous definition. Experiments have shown that accuracy and stability of optimization benefit from the consistency between forward and backward propagations of our differentiable renderer. Since we only focus on synthesizing silhouettes, only a few pixels and edges need to be considered. We employ quadtree to accelerate the process of retrieving edges which the gradient of current pixel may back-propagate into. As shown in the experiment, the efficiency of our implementation is higher than that of N3MR~\cite{kato2018neural}.

There are two main limitations of our method. One is that our differentiable renderer is not general-purpose which means that our method can not obtain derivatives with respect to texture and lighting parameters and limits the application in inverse graphic. The other is that our implementation requires constructing a quadtree recursively which leads to lower efficiency compared with previous method when the mesh is too simple or the resolution of output image is quite low.

Future direction of this work may include deriving analytical derivatives for general-purpose renderer to enable the gradients back-propagate into arbitrary scene parameters. It may also include developing a parallelizable algorithm to enable efficient implementation on GPU.

\bibliographystyle{splncs}
\bibliography{egbib}
\end{document}